\documentclass{article}
\usepackage{spconf,amsmath,graphicx}
\usepackage{multirow}
\usepackage{url}

\title{Domain-wise Invariant Learning for Panoptic Scene Graph Generation}
\name{Li Li$^{\dagger}$ \qquad You Qin$^{\dagger}$ \qquad Wei Ji$^{\dagger}$ \qquad Yuxiao Zhou$^{\dagger}$ \qquad Roger Zimmermann$^{\dagger}$}

\address{$^{\dagger}$National University of Singapore}
\begin{document}
%
\maketitle
\begin{abstract}
Panoptic Scene Graph Generation (PSG) involves the detection of objects and the prediction of their corresponding relationships (predicates). However, the presence of biased predicate annotations poses a significant challenge for PSG models, as it hinders their ability to establish a clear decision boundary among different predicates. This issue substantially impedes the practical utility and real-world applicability of PSG models.
To address the intrinsic bias above, we propose a novel framework to infer potentially biased annotations by measuring the predicate prediction risks within each subject-object pair (domain), and adaptively transfer the biased annotations to consistent ones by learning invariant predicate representation embeddings.
Experiments show that our method significantly improves the performance of benchmark models, achieving a new state-of-the-art performance, and shows great generalization and effectiveness on PSG dataset.
\end{abstract}
\section{Introduction}
\label{sec:intro}
Panoptic Scene Graph Generation (PSG) \cite{yang2022psg} aims to simultaneously detect instances and their relationships within visual scenes \cite{Chang_2023}. Instead of coarse bounding boxes used in Scene Graph Generation (SGG) \cite{10.1145/3474085.3475540,Xu_2017_CVPR,Lin_2020_CVPR,Li_2021_CVPR,pmlr-v119-chen20j,yao2022pevl,https://doi.org/10.48550/arxiv.2109.11797}, PSG proposed to construct more comprehensive scene graphs with panoptic segmentation \cite{Kirillov_2019_CVPR}. 

However, the current performance of PSG methods is suboptimal due to the biased prediction problem. The problem mentioned stems from the following two aspects:
(1) \textbf{Contradictory Mapping:} PSG models map visual instances to subjects/objects, and their relationships to predicates. However, annotators assign different predicate labels to identical subject-object pairs with similar image features due to their personal language preferences and the semantic ambiguity between predicates, leading to contradictory mapping from visual to linguistics.
(2) \textbf{Long-tail Distribution:} Existing models seriously entangle predicate prediction with the long-tail data distribution in the training dataset. Specifically, as long as the labels of the subject and object are known, the model can make effective predicate predictions even without resorting to any visual contents of an image \cite{9969654}.

Previous works \cite{Zellers_2018_CVPR,Yu2020CogTreeCT,yao2022pevl,Xu_2017_CVPR,Tang_2019_CVPR,Tang_2020_CVPR,Lin_2020_CVPR,li2023panoptic} exploit numerous model architectures to alleviate the bias problem, but these models achieve relatively limited performances, and cannot fundamentally solve the problem. \cite{zhang2022fine} have proposed to enhance the training dataset by a data transfer framework. However, their framework inaccurately transfers a significant number of samples, leading to imbalanced performance among predicates.

To alleviate the biased annotation problem, we propose constructing an unbiased dataset by transferring the biased annotations to high-quality consistent predicate annotations. 
Inspired by \cite{arjovsky2020invariant,LiMM}, we propose our framework which learns unbiased predicate representations excluding the influence from long-tailed subject-object pairs, for biased predicate annotations identification and transfer.
In the target inference process, we denote different subject-object pairs as different domains, and we measure the risk within each domain. Targets are then derived that partition the mapping of the reference model which maximally violates the ground truth labels.
In the invariant learning process, our aim is to exclude the spurious correlation between predicates and subject-object pairs. 
Specifically, we propose a predicate-wise invariant risk minimization method to learn invariant predicate representations without the influence from subject-object pairs. 
Meanwhile, we screen out potentially biased data by measuring their invariances within the dataset, to promise unbiased and invariant predicate representations.
Finally, with the unbiased predicate representation embedding space, biased annotations are easily transferred.

\begin{figure*}
\includegraphics[width=0.99\textwidth]{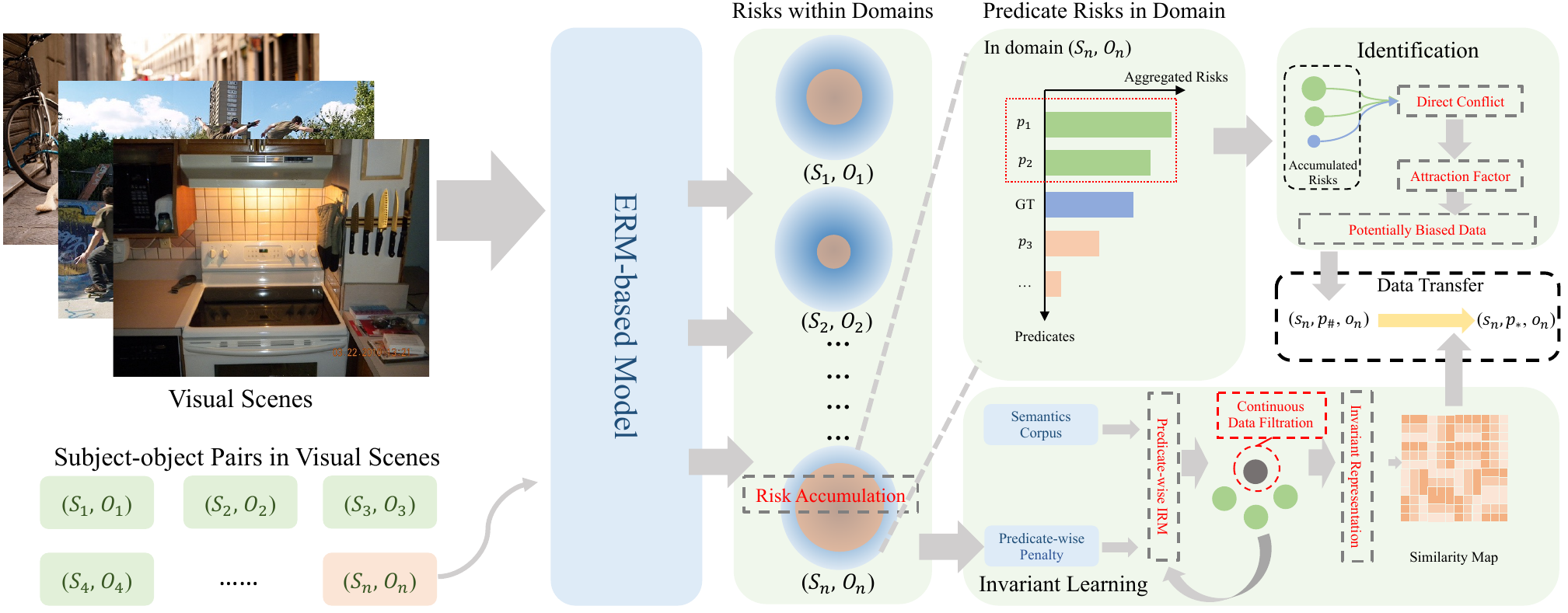} 
 \caption{\textbf{Illustration of the overall pipeline of our method.} It measures the ERM-based risks within domains to identify potentially biased data, and learns invariant predicate representations to make consistent data transfer.}
 \label{model structure}
\vspace{-1.5em}
\end{figure*}

In summary, the following contributions are made: (1) A novel, plug-and-play framework is proposed, which aims at adaptively and accurately performing biased-data transfer to promise a reasonable dataset with informative and standard-unified labels. (2) We propose a new domain-based invariant learning method, aiming at accurately identifying biased annotations, and promising consistency during the data transfer process. (3) Comprehensive experiments demonstrate that the proposed method significantly enhances the performance of benchmark models on the PSG dataset and achieves a new state-of-the-art performance.

\section{Method}
\subsection{Target Inference}
\label{TI}
We first measure the risks of ERM-based model assigned predicates within each domain, and then locate the assigned predicates that maximally violate the ground truth labels as the potentially biased predicate annotations.

\noindent \textbf{Risk within Domains.} 
Unbiased PSG methods aim to learn visual features that are predicate-invariant. However, due to the biased annotation and long-tail distribution problems, their predictions are seriously entangled with subject-object pairs (different domains). Thus, this step measure the risks of predicate predictions within each domain to help locate biased annotations.

We take the advantage of a reference classifier $  \tilde{\Phi}  $, which maps visual scenes and subject-object pairs (Input X) to predicates (Output Y) and is optimised with ERM on $ p^{obs}(X,Y) $. 
We begin by noting that the per-domain risk $ R^{d} $ depends implicitly on the manual subject-object labels from the training dataset. For a given domain $ d^{'} $, we denote $ I( d^{p} = d^{'} ) $ as an indicator that predicate $ p $ is assigned to that domain, and express the risk of predicate $ p $ in domain $ d $ as:

\begin{equation}
\label{RWD}
    R_{p}^{d^{'}}(\tilde{\Phi },X_{i} )={\textstyle\sum_{i}^{N} I(d^{p} = d^{'}) M(\tilde{\Phi}(X_{i}) ) },
\end{equation}
where $ M( \cdot ) $ denotes the mapping from visual scenes to classification distribution of predicate $ p $ of the ERM-based model, $ N $ denotes the number of samples in the training dataset with domain $ d^{'} $. In practice, we further normalize the risks of different predicates in a certain domain with softmax.

\noindent \textbf{Target Identification.}
With the risks of all predicates in each of the domain, we locate potentially biased predicate annotations by checking the mapped predicate classification distribution of the ERM-based model and the conflicts between the distribution and the ground truth.

The mapping $ M(\cdot) $ to predicate classification distribution provides a good reflection on the model's confusion on predicate prediction \cite{zhang2022fine}. Specifically, given a sample with subject-object pair (domain) $ d_{s} $ and predicted predicate $ p_{s} $,  we measure its Direct Conflict (DC) with the ground truth predicate by comparing the risks between $ p_{s} $ and the $ GT $ in the domain $ d_{s} $:

\begin{equation}
\label{DC}
    DC_{s}=R_{p_{s}}^{d^{s}}(\tilde{\Phi },X_{i} )-R_{GT}^{d^{s}}(\tilde{\Phi } ,X_{i}).
\end{equation}

The set of potentially biased annotations (target) $ S_{b} $ are located with the help of Direct Conflict as follows:

\begin{equation}
    S_{b}=\left \{ s_{I}|(DC_{s}>0)\wedge (A_{s}<A_{GT}) \right \},
\end{equation}
where fuction $ A $ denotes the attraction factor \cite{zhang2022fine} representing the scarcity of the predicate and domain.

\subsection{Invariant Learning}
\label{IL}
In this process, we aim to learn invariant predicate representations excluding the influence from subject-object pairs (domains).

\noindent \textbf{Predicate-wise Invariant Risk Minimization.}
 We first collect all of the annotations that appear in the training set.
Each annotation will be converted to a sentence for language model processing.
For example, $\textless$ person, standing on, road $\textgreater$ will be converted to \textit{The person is standing on the road}.

Formally, given an anchor sentence $ s_{i} $ in domain $ d_{i} $ of predicate class $ p_{i} $ in the batch $S=\left\{s_k\right\}_{i=k}^N$, we can construct its positive set ${S}_{i}^{+}=\{s_k|(p_i=p_k) \wedge (d_i=d_k) \}_{k\neq i}$ and negative set ${S}_{i}^{-}=\{s_k|(p_i \neq p_k) \wedge (d_i=d_k) \}_{k\neq i}$.
With the training data, our loss is computed within each domain $ d \in \varepsilon_{d} $, and we have:

\begin{equation}
\label{eq:rct}
\ell (d \in \varepsilon_{d})=\sum_{s \in d}\frac{1}{N^{+}}\sum_{z^{+}\in d}  -\log \frac{f^{+}(z^{T}z^{+})}{f^{+}(z^{T}z^{+})+f^{-}(z^{T}z^{-})},
\end{equation}
where $ N^{+} $ denotes the number of the positive samples in the current batch. Therefore, the proposed predicate-wise IRM loss is:

\begin{equation}
\label{IRM}
    L_{s}= {\textstyle \sum_{d\in\varepsilon _{d}}} \ell (d )+\lambda Var(\ell (d )),
\end{equation}
where $ Var(\cdot) $ denotes the variance of contrastive loss within each predicate class.

To further boost the sensitivity to predicate similarity, we introduce an angular margin $m$ for positive pairs.
Formally, we formulate the $ f^{+} $ as:
\begin{equation}
f^{+} = \sum_{s_{j} \in S_{i}^{+}} e^{cos ( \theta _{h_{i},h_{j}}+m )/\mathrm{T}},
\end{equation}
where $ \theta _{i,j} $ is the arc-cosine similarity between feature $ i $ and $ j $, $\mathrm{T}$ is a temperature hyper-parameter, \textit{N} is batch size, $h_{i,j}$ are language model generated sentence representations for $s_{i,j}$, and $ m $ is an angular margin introduced for robust learning.



For the samples in the negative set $ S^{-} $, we expect them to be quite different from the samples in the $ S^{+} $ in the embedding space. Thus, we propose a representation learning penalty for samples in the $ S^{-} $. Specifically, we calculate the predicate-wise risks taking advantage of Eq.~\ref{RWD}:

\begin{equation}
    \phi_{p} =\sum_{i}^{N} \sum_{d^{'} \in \varepsilon _{k}}R_{p}^{d^{'}}(\tilde{\Phi },X_{i} ).
\end{equation}

Then we measure the risk of negative sample with the predicate $ p_{k} $ for the anchor sample $ s_{i} $ with predicate $ p_{i} $:

\begin{equation}
\label{PIK}
    \varphi_{p_{i,k}} = Sigmoid(\phi_{p_{k}}-\phi_{p_{i}}).
\end{equation}

\begin{table}[]
\label{main}
\tabcolsep=0.09cm
\caption{The results (mR@K and PR@K) on SGDet task of our method and other baselines on PSG dataset. IETrans \cite{zhang2022fine} and Ours denote models equipped with different dataset-enhancement methods.}
\begin{tabular}{l|cccccc}
\hline
\multicolumn{1}{c|}{\multirow{2}{*}{Method}} & \multicolumn{6}{c}{Scene Graph Generation}                                                                         \\ \cline{2-7} 
\multicolumn{1}{c|}{}                        & mR@20         & @50           & \multicolumn{1}{c|}{@100}          & PR@20         & @50           & @100          \\ \hline
IMP \cite{Xu_2017_CVPR}                                         & 6.52          & 7.05          & \multicolumn{1}{c|}{7.23}          & 12.9          & 13.7          & 13.9          \\
\quad+IETrans                                     & 10.2          & 11.0          & \multicolumn{1}{c|}{11.3}          & 14.5          & 15.4          & 15.7          \\
\quad+Ours                                        &  \textbf{12.5}             &    \textbf{13.5}           & \multicolumn{1}{c|}{\textbf{14.0}}              &       \textbf{16.0}        & \textbf{17.1}              & \textbf{17.5}              \\ \hline
VCTree \cite{Tang_2019_CVPR}                                      & 9.70          & 10.2          & \multicolumn{1}{c|}{10.2}          & 16.0          & 16.8          & 16.9          \\
\quad+IETrans                                     & 17.1          & 18.0          & \multicolumn{1}{c|}{18.1}          & 19.6          & 20.5          & 20.7          \\
\quad+Ours                                        & \textbf{18.3} & \textbf{18.9} & \multicolumn{1}{c|}{\textbf{19.0}} & \textbf{20.3} & \textbf{20.8} & \textbf{20.9} \\ \hline
MOTIFS \cite{Zellers_2018_CVPR}                                      & 9.10          & 9.57          & \multicolumn{1}{c|}{9.69}          & 15.5          & 16.3          & 16.5          \\
\quad+IETrans                                     & 15.3          & 16.5          & \multicolumn{1}{c|}{16.7}          & 18.2          & 19.4          & 19.7          \\
\quad+Ours                                        & \textbf{18.4} & \textbf{19.0} & \multicolumn{1}{c|}{\textbf{19.2}} & \textbf{20.0} & \textbf{20.8} & \textbf{21.0} \\ \hline
GPSNet \cite{Lin_2020_CVPR}                                      & 7.03          & 7.49          & \multicolumn{1}{c|}{7.67}          & 13.6          & 14.4          & 14.7          \\
\quad+IETrans                                     & 11.5          & 12.3          & \multicolumn{1}{c|}{12.4}          & 15.3          & 16.2          & 16.5          \\
\quad+Ours                                        & \textbf{17.5} & \textbf{18.1} & \multicolumn{1}{c|}{\textbf{18.4}} & \textbf{19.6} & \textbf{20.3} & \textbf{20.6} \\ \hline
PSGTR \cite{yang2022psg}                                       & 16.6          & 20.8          & \multicolumn{1}{c|}{22.1}          & 21.9          & 26.3          & 27.6          \\
\quad+IETrans                                     & 23.1          & 27.2          & \multicolumn{1}{c|}{27.5}          & 24.9          & 28.4          & 28.7          \\
\quad+Ours                                        & \textbf{26.4} & \textbf{29.6} & \multicolumn{1}{c|}{\textbf{30.2}} & \textbf{27.0} & \textbf{29.9} & \textbf{30.4} \\ \hline
\end{tabular}
\vspace{-1.5em}
\end{table}

We treat the risk in Eq.~\ref{PIK} as the reflection of the visual similarity of predicates: Higher risk denotes harder prediction from the model. Sourcing back to the input of the model, it is the highly similar visual scenes. As a result, we use the metric $ (1- \varphi) $ to further differentiate similar predicate representations from the negative set.
Formally, we introduce the $ f^{-} $ as follows:

\begin{equation}
    f^{-} = \sum_{s_{g} \in S_{i}^{-}} \left (1-\varphi_{p_{i,g}}   \right )e^{\cos\left(\theta_{i,g}\right) / \mathrm{T}}.
\end{equation}

\begin{figure*}[!t]
\centering   \includegraphics[width=0.97\textwidth]{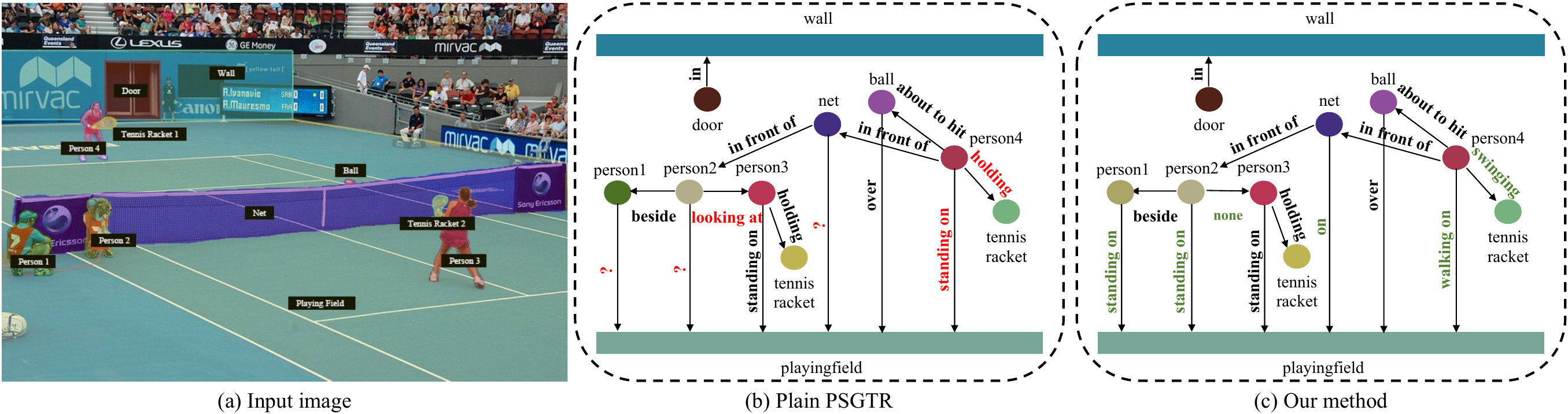}
    \caption{Visualization of plain PSGTR and PSGTR equipped with our method. PSGTR with our method can predict relationships between instances with greater accuracy and also select predicates that better match the visual scene.}
    \label{fig:performance}
\vspace{-1.5em}
\end{figure*}

\noindent{\textbf{Continuous Data Filtration.}}
\label{MDC}
The presence of biased and noisy samples within the training dataset is poised to exert a discernible impact on the impartial process of predicate representation learning.  Consequently, we have devised a continuous data filtration procedure to eliminate these biased samples. This approach leverages invariant representation regularization as a means of assessing the quality of samples.

We collect and average the variances from Eq.~\ref{IRM} on predicate labels, getting $ V_{aver}\in \mathbf{R}^{Q} $, where $ Q $ denotes the pre-defined predicate classes in the dataset. For every sample $ s_i $ with predicate label $ p_i $ and variance $ V_i $ in the training dataset, we judge whether it is part of potentially biased and noisy samples, which can be formulated as:

\begin{equation}
    P_{bn}=\left \{ S_i | V_i>\mu V_{aver}^{i}    \right \}, 
\end{equation}

where $ V_{aver}^{i} $ represents the averaged variance associated with predicate label $ p_i $, and $ \mu $ is a hyper-parameter. To further refine the dataset, we proceed to arrange the elements in $ P_{bn} $ in ascending order based on the loss value computed from Eq.~\ref{eq:rct}, subsequently excluding the uppermost $ D\% $ of the training data. It is noteworthy that, in cases where a predicate class contains fewer than 100 samples, no further elimination of samples is performed.

\subsection{Data Transfer}
\label{DT}
As a result, a similarity matrix $S\in\mathbf{R}^{Q\times Q}$ can be generated by calculating the cosine similarities between all predicate representations.  
The transfer method is based on importance vector \cite{zhang2022fine}.
We transfer the biased predicate annotation to a target predicate that shares the highest similarity and importance vector, and we directly use the similarity score as an adaptive transfer ratio.

\section{Experiment}
\label{sec:typestyle}
\subsection{Experiment Details}
\noindent\textbf{Dataset}. We evaluate our method on the PSG dataset \cite{yang2022psg}.

\noindent\textbf{Evaluation Metric.} Following previous works \cite{Tang_2020_CVPR,Zellers_2018_CVPR}, we take mean recall@K (mR@K)\cite{Tang_2019_CVPR,Chen_2019_CVPR} as evaluation metrics. Following PSG challenge \cite{yang2022psg}, we also adopt a new evaluation metric named percentile recall (PR), which can be formulated as $PR=30\%R+60\%mR+10\%PQ$, where PQ measures the quality of a predicted panoptic segmentation relative to the ground truth \cite{Kirillov_2019_CVPR,ji2023complexquery}.

\noindent{\textbf{Tasks.}} We evaluate our method on the Scene Graph Generation (SGDET) task.

\noindent\textbf{Implementation Details}. We use a {BERT-base} \cite{devlin-etal-2019-bert} for representation learning. The decision margin \textit{m} is set to 10 degrees, the temperature hyper-parameter T is set to 0.05, and we use an AdamW \cite{Loshchilov} optimizer with a learning rate 2e-5. The hyper-parameter $ \lambda $ is set to 0.3, $ \beta $ is set to 5e5, $ \mu $ is set to 1.2, and $ \gamma $ is set to 1.5.  The $ D $ in data filtration is set to 50. 

\subsection{Qualitative Analysis}
As depicted in Fig.~\ref{fig:performance}, a comparative analysis is conducted between the outcomes produced by the Plain PSGTR model and PSGTR integrated with our proposed method. Evidently, PSGTR augmented by our method exhibits an enhanced capacity for predicting more precise associations among instances, concurrently demonstrating a heightened ability to forecast predicates that align more fittingly with the contextual scene.

\subsection{Comparison with State-of-the-Art Methods}
Based on the obtained results in Tab.1, our proposed methodology significantly enhances the performance of baseline models across a wide spectrum of evaluation metrics. A comprehensive comparative analysis against the IETrans framework \cite{zhang2022fine} reveals notable advancements in terms of both mean recall and PR metrics across all baseline models. This observation underscores the superior efficacy of our methodology in optimizing the training dataset, effectively mitigating issues related to extraneous or duplicative transfer processes. Particularly noteworthy is the PR metric, which amalgamates considerations of recall and mean recall, showcasing our methodology's substantial superiority over the original models across a diverse range of predicate labels. This affirms that our approach not only ameliorates recall performance but also adeptly balances the overall performance across various predicate labels, resulting in a more comprehensive and robust assessment of model performance.

\begin{table}[]
\label{DPM}
\tabcolsep=0.16cm
\caption{Ablation study on data processing methods. Transfer: data transfer. Remove: simply remove all identified biased data. Original: baseline method on the original dataset.}
\begin{tabular}{c|ccc}
\hline
\multirow{2}{*}{Data Processing Method} & \multicolumn{3}{c}{SGDet}               \\ \cline{2-4} 
                                 & mR@20     & mR@50     & mR@100    \\ \hline
Original                         & 16.6 & 20.8 & 22.1 \\ \hline

Remove                           & 20.0 & 24.6 & 25.3 \\ \hline

Transfer                 & 26.4 & 29.6 & 30.2 \\ \hline
\end{tabular}
\vspace{-1.8em}
\end{table}


\subsection{Ablation Study}
We use PSGTR as the baseline model in ablation studies.
Despite the data transfer method, we directly remove the potentially biased data to prove the effectiveness of our target inference process, and to prove the harm of these biased data in training process. As shown in Tab.2, baseline model easily achieves great performance with only biased data removed, and its performance can be further enhanced with our data transfer method.



\section{Conclusion}
\label{sec:majhead}
We present a novel framework for PSG aimed at mitigating the issue of biased prediction. This framework identifies and transfers biased annotations, ensuring a more balanced and representative training dataset. Empirical findings substantiate the performance improvements achieved by our method, consequently establishing a new state-of-the-art benchmark.

\section{Acknowledgement}
This research is supported by Singapore Ministry of Education Academic. Research Fund Tier 2 under MOE's official grant number T2EP20221-0023.

\bibliographystyle{IEEEbib}
\bibliography{strings,refs}

\end{document}